\documentclass[11pt,a4paper]{article}
\usepackage[hyperref]{acl2021}
\usepackage{times}
\usepackage{latexsym}

\usepackage{microtype}

\usepackage{graphicx}
\usepackage[T1]{fontenc}
\usepackage{float}
\usepackage{enumitem}
\usepackage{booktabs}
\usepackage{dblfloatfix}

\aclfinalcopy

\newcommand{\corpus}{\texttt{ding-01}}

\title{\corpus{} \texttt{:ARG0}: An AMR Corpus for Spontaneous French Dialogue}

\author{Jeongwoo Kang\textsuperscript{$\forall$} \quad Maria Boritchev\textsuperscript{$\exists$} \quad Maximin Coavoux\textsuperscript{$\forall$}\\
 \textsuperscript{$\forall$} Univ. Grenoble Alpes, CNRS, Grenoble INP,  LIG, 38000 Grenoble, France \\ 
 \textsuperscript{$\exists$} LTCI, Télécom Paris, 91120 Palaiseau, France \\ 
 \texttt{jeongwoo.jay.kang@gmail.com} \\
 \texttt{maria.boritchev@telecom-paris.fr} \\
 \texttt{maximin.coavoux@univ-grenoble-alpes.fr}
}

\date{}

\begin{document}
\maketitle

\begin{abstract}
We present our work to build a French semantic corpus by annotating French dialogue in Abstract Meaning Representation (AMR).
Specifically, we annotate the DinG corpus, consisting of transcripts of spontaneous French dialogues recorded during the board game \textit{Catan}.
As AMR has insufficient coverage of the dynamics of spontaneous speech, we extend the framework to better represent spontaneous speech and sentence structures specific to French.
Additionally, to support consistent annotation, we provide an annotation guideline detailing these extensions. We publish our corpus under a free license (CC-SA-BY). We also train and evaluate an AMR parser on our data. This model can be used as an assistance annotation tool to provide initial annotations that can be refined by human annotators. 
Our work contributes to the development of semantic resources for French dialogue.
\end{abstract}

\section{Introduction}
\textit{Abstract Meaning Representation} \citep[AMR]{banarescu-etal-2013-abstract} encodes the meaning of a text as a rooted, directed, and acyclic graph (see Figure~\ref{fig:amr_graph}). Representing meaning in a structured form offers several advantages for information systems. AMR reduces semantic ambiguity by explicitly specifying one plausible interpretation among others. Furthermore, because AMR abstracts away from surface variations — especially syntactic variations — sentences with the same underlying meaning share the same AMR representation (\textit{e.g.}, ``The police arrested the thief.'' and ``The thief was arrested by the police.''). This canonical representation reduces the search space for models, making AMR a useful tool for various NLP tasks, such as machine translation \citep{wein-schneider-2024-lost}, automatic text summarization \citep{liao-etal-2018-abstract, liu-etal-2015-toward}, and human-robot interaction \citep{bonial-etal-2019-abstract, bonial-etal-2023-abstract}.

Training an AMR parser to automatically generate an AMR graph from a given text requires a dataset consisting of texts associated with their corresponding AMR graphs.
However, AMR datasets for French are currently scarce, since most available AMR resources are in English. This imbalance in semantic resources limits the development of French semantic parsers, which hinders the progress of French NLP systems that rely on them. Furthermore, most existing AMR data are based on written texts such as newspaper articles and online forums.
In contrast, dialogue data, which exhibits unique linguistic features due to its interactive and spontaneous nature --\textit{e.g.,} French discourse markers such as \textit{alors} (then), \textit{du coup} (so), \textit{donc} (so), and backchannels-- remain underrepresented. 

To fill this gap in French semantic resources, particularly for dialogue, we manually annotate the DinG corpus \citep{boritchev-amblard-2022-multi} in AMR. DinG consists of transcriptions of dialogues recorded during board game sessions of \textit{Catan}, capturing various linguistic features of spoken interaction in French.

However, the standard AMR framework, as currently defined,\footnote{The current version of the annotation guideline is available at \url{https://github.com/amrisi/amr-guidelines/blob/master/amr.md}}
has limitations in representing speech-specific features. Therefore, we extend AMR by introducing additional relations to (i) annotate two pragmatic phenomena: discourse markers and backchannel expressions, (ii) represent coreference across multiple turns of speech. 

To summarize, our main contributions are as follows:
\begin{itemize}
\item We publish \corpus{},\footnote{\url{ https://doi.org/10.5281/zenodo.15537425}}
%\footnote{\url{https://propbank.github.io/v3.4.0/frames/ding.html\#ding.01}}}
a new AMR corpus of spontaneous French dialogue containing 1,830 turns of speech. We aim to expand the corpus to cover 3,000 turns of speech by the end of 2025.
We also release a \textit{data statement} with the corpus to describe all relevant metadata and potential biases, following best practices for data production for NLP \citep{bender-friedman-2018-data,10.1145/3594737}.
\item We adapt AMR to represent spontaneous speech phenomena in French, including discourse markers and backchannels.
\item We provide an annotation guideline for two purposes: 1) ensure annotation consistency by clarifying aspects not specified in the original AMR annotation guideline 2) newly define how to annotate linguistic features specific to French dialogue.
\item We train and evaluate an AMR parser on our dataset to showcase its practical use case. This model is further expected to serve as an annotation assitance tool. 
\end{itemize}

We expect our corpus to contribute to the future development of semantic parsers for French dialogue, along with future (computational) linguistics research on French dialogical data. %\citet{wein2024survey} list linguistics research using AMR as an underexplored usage of AMR so far, thus we produced a corpus and a parser to provide tools for this research direction.
As noted by \citet{wein2024survey}, AMR corpora and tools are an underexplored source of data for linguistic investigation.
The corpus is already getting some interest from the semantics research community, as it has been integrated in Grew \citep{amblard2022graph} and can now be explored in the tool.\footnote{\url{https://semantics.grew.fr/?corpus=ding-01}} 

\section{Background and Related Work}
\subsection{Introduction to AMR}
\begin{figure}[H]
\centering
  \includegraphics[width=.25\textwidth]{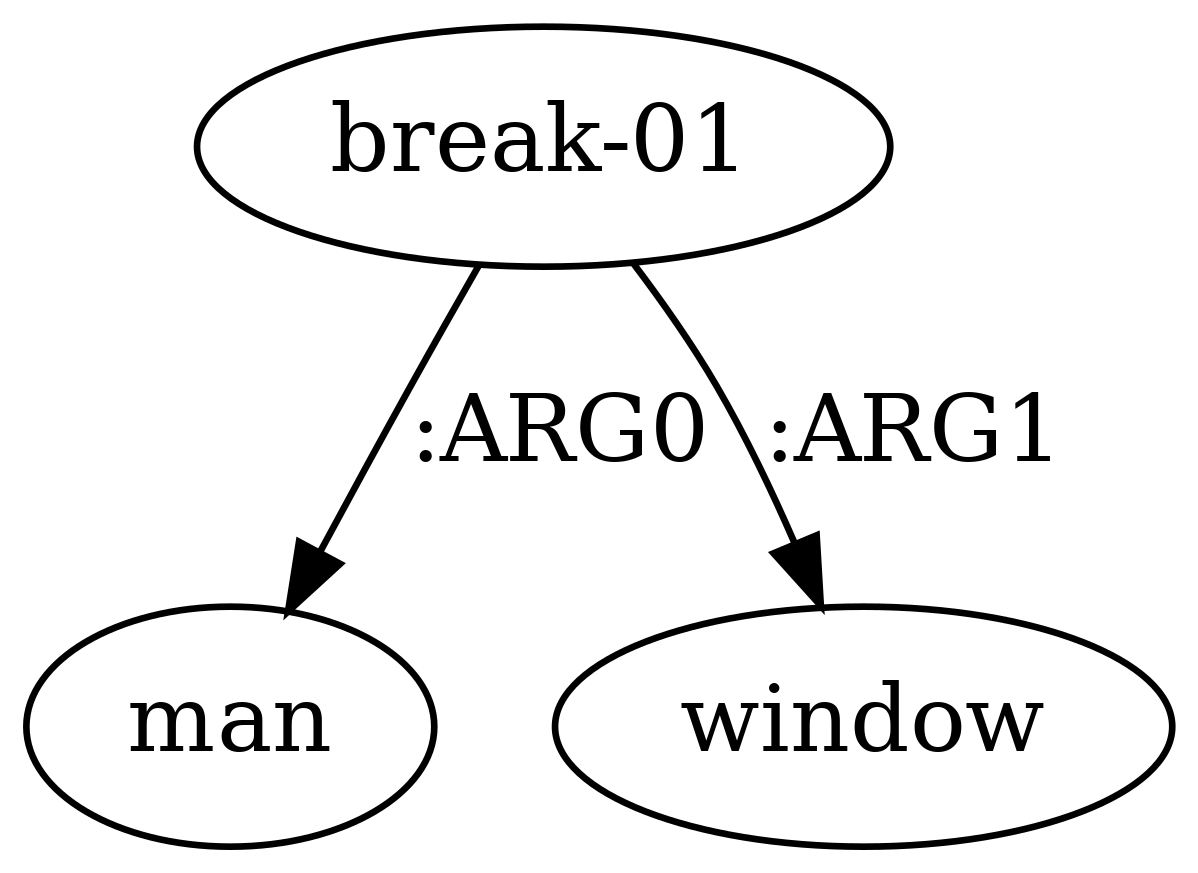}
  \caption{AMR graph for ``A man breaks a window'' or \guillemotleft~Un homme a cassé la fenêtre~\guillemotright.}
  \label{fig:amr_graph}
\hfill
\begin{minipage}{.48\textwidth}
  \centering
  \raggedright\vspace{13pt}
  {\small\texttt{\qquad\qquad(b / break-01 \\
  \qquad\qquad\qquad:ARG0 (m / man) \\
  \qquad\qquad\qquad:ARG1 (w / window)) \\}}
  \captionof{figure}{AMR graph linearized in text format.}
  \label{fig:amr_penman}
\end{minipage}
\end{figure}

AMR represents the meaning of texts using directed, acyclic, and rooted graphs. In an AMR graph, the nodes are 1) predicates predefined in Propbank\footnote{\url{https://propbank.github.io/v3.4.0/frames/}} \citep{palmer-etal-2005-proposition}, \textit{e.g.}, \texttt{break-01} in Figure~\ref{fig:amr_graph} or 2) English words, \textit{e.g.}, \texttt{man} and \texttt{window} in Figure~\ref{fig:amr_graph} or 3) AMR-specific keywords, \textit{e.g.}, \texttt{date-entity}.

The edges of the AMR graph are labeled to indicate the relation between nodes. For example, \texttt{:ARG0} and \texttt{:ARG1} in Figure~\ref{fig:amr_graph} respectively indicate that \texttt{man} is the agent of the predicate \texttt{break-01} and that \texttt{window} is the object of the same predicate. This predicate-argument structure is defined in Propbank.\footnote{\url{https://propbank.github.io/v3.4.0/frames/break.html\#break.01}}
An AMR graph can also be represented in textual form (see Figure~\ref{fig:amr_penman}). Although AMR is initially designed for English texts, it is also commonly used to represent non-English texts \citep{damonte-cohen-2018-cross,xu-etal-2021-xlpt,liu-etal-2020-multilingual-denoising}. In multilingual settings, two sentences in different languages that convey the same meaning (\textit{i.e.}, sentences that are translations of each other) will share the same AMR graph.

\subsection{AMR Datasets}
Most large-scale AMR datasets, including AMR~3.0 \citep{knight-etal-2020-ldc2020} and Massive-AMR \citep{regan-etal-2024-massive}, are available exclusively in English. AMR~3.0 is the most popular dataset for training and evaluating AMR parsers. It contains around 60,000 annotated examples from various sources such as news articles, blogs, and online forums. Massive-AMR, the largest manually annotated AMR dataset, consists of 84,000 utterances addressed to a virtual assistant. Most sentences in Massive-AMR are short questions or requests.

For French, a few datasets are available: \textit{Le Petit Prince} AMR \citep{kang-etal-2023-analyse}, Massive-AMR French \citep{regan-etal-2024-massive} and ReMEDIATE \citep{druart:tel-04963857}. For \textit{Le Petit Prince} AMR, the authors manually aligned the entire English dataset, \textit{The Little Prince} AMR,\footnote{\url{https://github.com/flipz357/AMR-World/blob/main/data/reference_amrs/amr-bank-struct-v3.0.txt}} with the original French text. The French Massive-AMR consists of a part of Massive-AMR English \citep{regan-etal-2024-massive}, manually translated into French. ReMEDIATES is annotated semi-automatically in French using a trained annotation model. Unlike two previous datasets, ReMEDIATES is not built on pre-existing English data. In terms of corpus type, \textit{The Little Prince} AMR is a literary piece of work. Massive-AMR consists of requests sent to virtual assistants. Finally, ReMEDIATES contains interactions between a virtual assistant and its user to make reservations. Note that ReMEDIATES uses the syntax of AMR graphs but adapts all the concepts and edge labels for Task-Oriented Dialogues (TOD).

Our work stands out from prior work in several key ways.
First, we annotate spontaneous conversations between multiple speakers.
Our corpus captures real-world interactions, reflecting the dynamics of spontaneous speech in French. Furthermore, \textit{The Little Prince} AMR and Massive-AMR were initially annotated in English and then adapted to other languages through manual translation or crosslingual alignment (assuming that translated sentences should have the same semantic graph as its original sentence). This process can introduce bias, making the data potentially English-centric. We directly annotate French dialogues in AMR without relying on prior English annotations, ensuring that the semantics of French are preserved throughout the annotation process.
Finally, while ReMEDIATES is annotated semi-automatically, we annotate the data manually. It is worth emphasizing that large generative language models remain unreliable for semantic annotation tasks, even for English \citep{ettinger-etal-2023-expert}.

\subsection{AMR for Dialogues}
Although standard AMR provides various semantic roles to present meanings of \textit{texts}, several efforts have been made to extend it to capture various aspects of \textit{dialogue}. DMR \citep{hu-etal-2022-dialogue} and Dialogue-AMR \citep{bonial-etal-2020-dialogue}, as well as the work of \citet{druart:tel-04963857} are among these extensions. These three approaches primarily focus on task-oriented dialogues, in which an agent requests an action to a robotic or virtual agent. Therefore, they integrate fine-grained instructions and introduce additional roles to represent, for example, illocutionary force or the speakers' intended contribution \citep{bonial-etal-2020-dialogue}.

However, these roles are not ideally suited to our corpus, which consists of spontaneous conversations among multiple speakers. While we aim to follow standard AMR conventions as closely as possible by adhering to the established annotation guidelines, the nature of our data—French dialogue—introduces linguistic phenomena specific to natural oral interaction, such as backchannels and discourse markers.

Backchannels and discourse markers convey pragmatic information in dialogue. However, standard AMR does not take this type of information into account, as specified in its annotation guidelines. Despite this, we chose to annotate the pragmatic information conveyed by backchannels and discourse markers for two main reasons. First, unlike AMR~3.0, which relies primarily on textual data, our corpus consists of dialogues rich in pragmatic content. We believe that annotating this information provides a valuable resource for the study of French dialogue. Furthermore, the additional roles we propose can be easily removed, ensuring compatibility with AMR~3.0.

Second, although the AMR annotation guideline states that pragmatic information is not included, in practice, AMR incorporates some pragmatic elements. For example, the choice of the root node in AMR often depends on the primary focus of the sentence, reflecting pragmatic information. In addition, some predicates (\textit{e.g.,} \texttt{know-05} and \texttt{see-03}) are used for their discourse functions (\textit{e.g.,} as in ``you know'' and ``you see.''), which are also closely related to pragmatics. Thus, adding pragmatic elements to our annotations is not entirely incompatible with standard AMR practices. To account for this pragmatic information, we introduce new roles, which are detailed in Section~\ref{sec:amr-ding}.

\section{The DinG Corpus}
We annotate the DinG corpus\footnote{\url{https://gitlab.inria.fr/semagramme-public-projects/resources/ding/}} \citep{boritchev-amblard-2022-multi}, a collection of manually transcribed multi-party dialogues among French-speaking players of the board game Catan.\footnote{We refer readers to the website \url{https://www.catan.com/} for more information on the game.} Catan is a strategic board game centered on resource management and exchange. Thus, players often negotiate resource exchanges with each other, and their actual interactions are recorded in the corpus. We select this corpus for two main reasons.

First, DinG is available under a free license.\footnote{The \textit{Attribution ShareAlike Creative Commons} (CC BY-SA~4.0) license.} As our goal is to make our data public, selecting open data is a crucial requirement. Second, DinG consists of natural dialogues among speakers. Since the environment is not controlled by the data collectors and the players are free to interact during the game, this dataset captures a natural conversational flow and includes a wide variety of dialogic phenomena. As such, its semantic annotations will serve as an ideal testbed for evaluating pre-trained language models on spontaneous speech transcriptions.

\begin{figure*}
    \centering
    \includegraphics[width=0.85\textwidth]{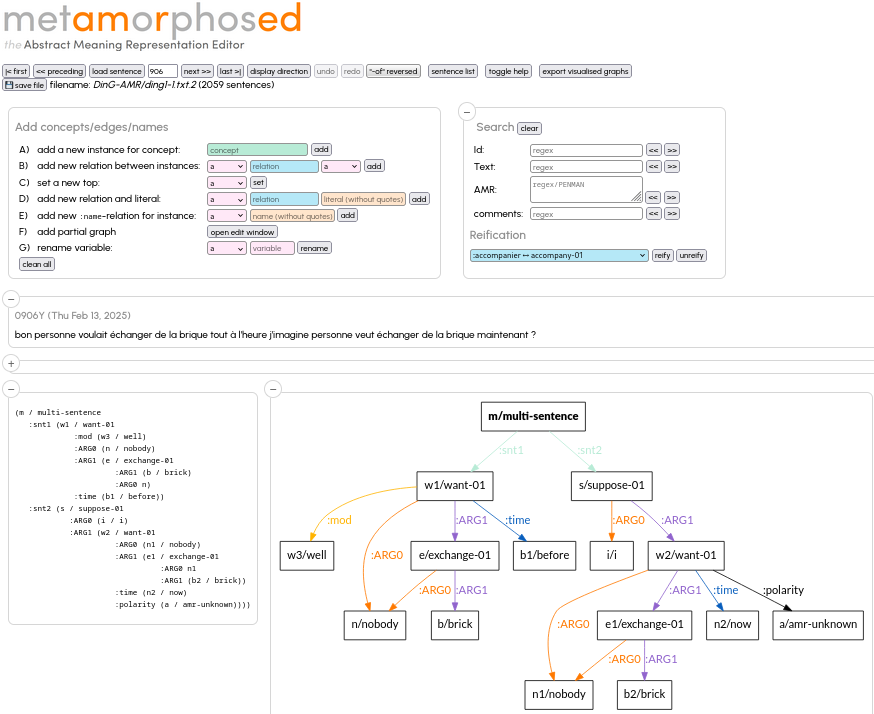}
    \caption{Screenshot illustrating the annotation process with metAMoRphosED.}
    \label{fig:metamorphosed}
\end{figure*}

\section{\corpus{}}
\label{sec:corpus}

In this section, we present some statistics on the corpus, the annotation process, and the data quality assessed by inter-annotator agreement.

The annotation was carried out over a six-months period, during which approximately~1,830 (see Table \ref{tab:stats} for other statistics) turns of speech were annotated using AMR.\footnote{We followed the original turn-taking divisions as defined in the DinG corpus.}
%\footnote{Among~1,830 turn takings annotated, some example only consist of non-annotable words, \textit{e.g.,} \textit{[toux]} (cough), \textit{[rire]} (laugh). The number of utterances (non-empty) excludes these non-annotable examples.}) turns of speech were annotated using AMR.\footnote{We followed the original turn-taking divisions as defined in the DinG corpus.}
Among these~1,830 turn takings, some examples only consist of non-annotable words, \textit{e.g.,} \textit{[toux]} (cough), \textit{[rire]} (laugh). 
The number of utterances (non-empty) in Table~\ref{tab:stats} excludes these non-annotable examples.

Among these examples, there are~459 discourse markers and~36 instances of \textit{backchannel}. The corpus was primarily annotated by the first author of this article using the \textit{metAMoRphosED} annotation tool \citep[][see Figure \ref{fig:metamorphosed}]{heinecke-boritchev-2023-metamorphosed}. Approximately~15\% of the examples in the entire corpus were validated by two other annotators, who are co-authors of this article. Specifically, the lead annotator and the two annotators met regularly throughout the annotation process (once a week or every two weeks) to check the validity of the examples one by one and record any difficulties encountered.
In case of disagreement among the three annotators, the example was corrected or modified during the discussion.

\begin{table}[]
\resizebox{\columnwidth}{!}{
\begin{tabular}{lr}
\toprule
%                     & Number  \\ \midrule
Number of utterances (non-empty) & 1,667  \\
Number of tokens covered & 17,887 \\
Number of speakers   & 9      \\ \bottomrule
\end{tabular}}
\caption{Basic statistics on our data.}
\label{tab:stats}
\end{table}

We encountered several challenges during the annotation process.
% (the annotation process is illustrated in figure~\ ref{fig:metamorphosed_annotation})
One example concerned the word `\textit{donc}' (so), which appears frequently in DinG. In most cases, it functions more as a discourse marker (used to start a speech turn or as a filler word) than as a causal connector. However, its usage was often ambiguous, and both interpretations could be valid depending on the context. To reduce ambiguity and improve consistency between annotations, we established the following rule: systematically annotate `\textit{donc}' as a discourse marker, provided that its removal does not change the meaning of the sentence. Our method for addressing other similar challenges by defining clear directions is detailed in our annotation guidelines. Furthermore, when faced with complex cases, or cases where multiple annotation choices were correct, we referred to existing AMR~3.0 data in English to choose the most plausible annotation.
These examples contain comments with references to the AMR~3.0 sentences that justify these choices.

To assess the quality of the annotations, 160 examples from our corpus were annotated by two annotators. The agreement score was measured using the \textsc{smatch} \citep{cai-knight-2013-smatch} score. \textsc{smatch} is an evaluation metric for AMR calculated by counting the number of triplets (node, labeled edge, node) in common. We obtained a score of~71.6. For comparison, \citet{banarescu-etal-2013-abstract} reports inter-annotator agreement scores ranging from 71 to 83, depending on the data source and the annotators' level of expertise.

After this evaluation, we performed an annotation conflict resolution step to produce our final \textit{gold} corpus. All three authors jointly reviewed these 160 annotation examples. In cases of disagreement, the group resolved conflicts by choosing one of the existing annotations or agreeing on a new alternative.

Common conflicts involved edge labels such as \texttt{:ARG0}, \texttt{:ARG1}, and \texttt{:ARG2}, typically resulting from annotation mistakes that were straightforward to correct once identified. Another recurring issue concerned the selection of synonymous PropBank concepts. For instance, \texttt{own-01} and \texttt{possess-01} convey the same meaning and share the same two semantic roles (\texttt{:ARG0} for the owner and \texttt{:ARG1} for the owned item). In the English AMR data, the choice between these concepts is guided by the specific lexical item used in the sentence. We used these cases of conflict to refine our annotation guidelines, ensuring a consistent selection between such synonymous concepts.

\section{AMR Adapted for DinG} \label{sec:amr-ding}

While adhering as closely as possible to standard AMR, we introduce some extensions to better capture the specific features of spontaneous French speech. Some of these key features are outlined below. In addition, we annotate inter-instance coreference, which is an addition that sets our corpus apart from AMR~3.0. We also adapt the standard AMR concept of \textit{focus} to represent focalization strategies in spoken French. 
Further details on these extensions are provided in our annotation guideline.

For \corpus{} use cases requiring compatibility with the English AMR~3.0 corpus, these extensions are designed to be easily removable. % from our data.

\subsection{Discourse Markers}

Discourse markers are short words or phrases used by speakers to structure their discourse, for example, \textit{donc} (so), \textit{et} (and). They are used to begin an utterance, or can serve as fillers in the middle of an utterance or during a hesitation. We introduce a new role, \texttt{:discourse-marker}, to annotate them (see Figure~\ref{fig:discourse_marker}).
This role can also be reified with the concept \texttt{be-discourse-marker-91}.

\begin{figure}[H]
\centering
\begin{minipage}{.45\textwidth}
  \centering
  \raggedright
  {\footnotesize\texttt{\#::id 0780B\\
                 (p / put-01\\
                 \qquad:ARG0 (y / you)\\
                 \qquad:ARG1 (r / road)\\
                 \qquad:mode imperative\\
                 \qquad:ARG2 (h / here)\\
                 \qquad:polarity -\\
                 \qquad:\textbf{discourse-marker ``donc''}}}\vspace{7pt}
  \captionof{figure}{« \textbf{Donc} mets pas ta route ici » (So don't put your road here).\protect\footnotemark}
  \label{fig:discourse_marker}
\end{minipage}%
\end{figure}

\footnotetext{\texttt{\#::id} specifies the identifier of the example in our corpus. The identifier is composed of a number (i.e., \texttt{0780}) and the letter (i.e., \texttt{B}) that denotes a speaker.} 

\subsection{Backchannels}
\textit{Backchannels} refer to short interjections made by a listener while another person is speaking (e.g., \textit{hum}, \textit{mmh-mmh}) to signal attention to the conversation. We annotate them using a new relation \texttt{:back-channel}, which can be reified  with the concept \texttt{be-back-channel-91}.  Figure~\ref{fig:backchannel} is an annotation of backchannel to a previous utterance (Figure~ \ref{fig:backchannel_context}).

\begin{figure}[H]
\centering
\begin{minipage}{.45\textwidth}
  \centering
  \raggedright
  {\scriptsize\texttt{\#::id 0851B\\
  (p / possible-01 \\
   \qquad:ARG1 (e / exchange-01 \\
   \qquad\qquad        :ARG1 (t / thingy)) \\
   \qquad:ARG1-of (r / request-confirmation-91) \\
   \qquad:discourse-marker ``du coup'' \\
   \qquad:time (n / now))
}}
  \captionof{figure}{« du coup là on peut échanger des trucs c'est ça ? » (So now we can exchange thingies, right?).}
  \label{fig:backchannel_context}
\end{minipage}
\hfill
\centering
\begin{minipage}{.45\textwidth}
  \centering
  \raggedright
  {\footnotesize\texttt{\hspace{1cm}\\\qquad \#::id 0852Y\\
  \qquad(b / \textbf{be-back-channel-91}\\
   \qquad\qquad:ARG2 ``hum'')\\}}
  \captionof{figure}{« hum » (hmm).}
  \label{fig:backchannel}
\end{minipage}
\end{figure}

\subsection{Inter-Instance Coreference}

Since the DinG corpus captures interactions between players throughout the game, coreference can span multiple utterances or instances. To ensure a complete representation of meaning, we annotate multi-instance coreferences by marking antecedents that appear in different utterances. For example, the node \texttt{s0080b\_s\_stone} in Figure~\ref{fig:coref2} indicates that its antecedent comes from the example identified by the ID \texttt{0080b} in Figure~\ref{fig:coref1} and the concept \texttt{s / stone} associated with that example.

\begin{figure}[H]
\centering
\begin{minipage}{.45\textwidth}
  \centering
  \raggedright
  {\footnotesize\texttt{\# ::id 0080B\\
            (w / want-01 \\
            \qquad:ARG0 (y / you) \\
            \qquad:ARG1 (\textbf{s / stone}) \\
            \qquad:polarity (a / amr-unknown)) \\}}\vspace{7pt}
  \captionof{figure}{« Tu veux de la \textbf{pierre} ? » (You want stone?) }
  \label{fig:coref1}
\end{minipage}%

\begin{minipage}{.45\textwidth}
  \centering
  \raggedright
  {\footnotesize\texttt{\qquad\# ::id 0082B \\
  \qquad(e / exchange-01 \\
   \qquad\qquad:ARG0 (I / I) \\
   \qquad\qquad:ARG2 (y / you) \\
   \qquad\qquad:ARG1 (s / sheep \\ 
   \qquad\qquad\qquad:quant 3) \\ 
   \qquad\qquad:ARG3 (s1 / \textbf{s0080B\_s\_stone}))\\}}
  \captionof{figure}{« Je te \textbf{l}'échange contre 3 moutons » (I trade you 3 sheep for it).}
  \label{fig:coref2}
\end{minipage}
\end{figure}

\subsection{Inter-Instance Verb Ellipsis}

Speakers often omit verbs when the meaning remains clear without them (verb ellipsis). When this occurs across different instances (inter-instance level), the omitted verb is mentioned in a previous utterance, and may be spoken by another speaker. 
We annotate such ellipses similarly to inter-instance coreference, by referencing the utterance ID of the original verb (see Figure~\ref{fig:ellipsis1} and~\ref{fig:ellipsis2}). 

\begin{figure}[H]
\centering
\begin{minipage}{.45\textwidth}
  \centering
  \raggedright
  {\footnotesize\texttt{\# ::id 0061R\\
            (a / and \\
            \qquad:op2 (p / possible-01 \\
            \qquad\qquad:ARG1 (\textbf{p1 / put-01} \\
            \qquad\qquad\qquad\qquad:ARG0 (\textbf{w / we)} \\
            \qquad\qquad\qquad\qquad:ARG1 (c / settlement) \\
            \qquad\qquad\qquad\qquad:ARG2 (i / intersect-01)  \\
            \qquad\qquad\qquad\qquad:mod (o / only))))\\}} 
  \captionof{figure}{« On peut poser les colonies que sur les intersections. »  (We can put the settlements only on intersections).}
  \label{fig:ellipsis1}
\end{minipage}%
\hfill
\begin{minipage}{.45\textwidth}
  \centering
  \raggedright\vspace{11pt}
  {\footnotesize\texttt{\qquad\# ::id 0062Y \\
  \qquad(s / \textbf{s0061R\_p1\_put-01} \\
   \qquad\qquad:ARG0 (w / \textbf{s0061R\_w\_we}) \\ 
   \qquad\qquad:ARG1 (r / road) \\
   \qquad\qquad:ARG2 (e / edge \\
   \qquad\qquad\qquad:mod (o / only))\\}}
  \captionof{figure}{« et les routes que sur les arêtes » (and roads only on edges).}
  \label{fig:ellipsis2}
\end{minipage}
\end{figure}

\subsection{Focus Representations}
In AMR, the \textit{focus} of a sentence is indicated by a root node.
We apply this principle to the annotation of cleft structure, a sentence structure commonly used in French for emphasis. The cleft structure follows the pattern « C'est [subject] qui ... » (``it's [subject] who/that...'' in English) used to emphasize the [subject]. To reflect this emphasis on the subject, we select it as the root of the AMR graph. Figure~\ref{fig:structure_clive} presents an example of a sentence with a cleft structure, accompanied by its annotation in AMR.
We adopt the same strategy for cases of left dislocations with pronominal resumption, as in the example: «moi, je veux 2 blés» (``me, I want 2 grains,'' in English). This type of structure, very common in spoken French, is also a way of expressing focus. In this case, the concept \texttt{i} will be the root of the AMR graph.

\begin{figure}[H]
\centering
\begin{minipage}{.45\textwidth}
  \centering
  \raggedright
  {\footnotesize\texttt{\#::id 0095Y\\
  (\textbf{y / you}\\
   \qquad:ARG0-of (c / choose-01 \\
    \qquad\qquad:ARG1 (p1 / place \\
    \qquad\qquad\qquad:ARG2-of (p / put-01 \\
    \qquad\qquad\qquad\qquad:ARG1 (t / they))))\\
    \qquad:polarity (a / amr-unknown))\\}}\vspace{11pt}
  \captionof{figure}{« \textbf{C'est toi qui} choisis où est-ce que tu les mets ? » (It's \textbf{you} who choose where you put them?).}
  \label{fig:structure_clive}
\end{minipage}%
\end{figure}

\subsection{Disfluencies}
Disfluencies are common in spontaneous dialogues. Disfluency markers (\textit{e.g.}, euh, eh), repetitions (\textit{e.g.},\ «franchement t'es t'es franchement» ``frankly you're you're frankly'' in English) and false starts (\textit{e.g.},\ «j'ai be- j'ai pas de bois» ``I nee- I don't have lumber'' in English) are often observed in the DinG corpus. In standard AMR, disfluency markers are not annotated. In line with this convention, we do not annotate disfluency markers, repetitions or short false starts. However, if a false start has interpretable semantic content, we annotate it using \texttt{:reparandum} (see Figure \ref{fig:reparandum}) following \citet{de-marneffe-etal-2021-universal}, who employed this label to mark overridden disfluencies in syntactic annotations.

\begin{figure}[H]
\centering
\begin{minipage}{.45\textwidth}
  \centering
  \raggedright
  {\scriptsize\texttt{\# ::id 0314R\\
  (t / thing \\
   \qquad:value 7 \\
   \qquad:ord (o / ordinal-entity \\
    \qquad\qquad:value 1) \\
   \qquad:ARG1-of (f / fall-01) \\
   \qquad:ARG1-of (h / have-degree-91 \\
    \qquad\qquad:ARG5 (r / roll-01 \\
    \qquad\qquad\qquad:ARG1 (d / dice)) \\
    \qquad\qquad:ARG2 (c / common \\
    \qquad\qquad\qquad\textbf{:reparandum (p / possible-01)}) \\
    \qquad\qquad:ARG3 (m / most)) \\
   \qquad:discourse-marker ``et'' \\
   \qquad:discourse-marker ``donc'' \\
   \qquad:discourse-marker ``hein'' \\ 
   \qquad:discourse-marker ``et'')
  }}
  \captionof{figure}{« et au premier 7 qui va tomber qui est donc euh le lancé de dés \textbf{le plus possible hein le plus courant} » (and the first 7 to fall, which is the most posssible the most common dice roll).}
  \label{fig:reparandum}
\end{minipage}
\end{figure}

\section{Models}
We train an AMR parser on the previously described data to showcase its practical use. The trained model can assist in the annotation process in our future work. Specifically, the model automatically annotates the data, which can then be manually refined by a human annotator. This semi-automatic approach is useful for scaling up data annotation.  

\subsection{Sequence-to-Sequence AMR Parser}

Recently, sequence-to-sequence AMR parsers \citep{konstas-etal-2017-neural,Bevilacqua_Blloshmi_Navigli_2021,yu-gildea-2022-sequence} have gained popularity due to their strong performance and methodological simplicity. These models take an input sentence and generate an AMR graph in a textual format. Training such models requires a graph linearization step, which converts the AMR graph into a single-line textual format. It also requires a post-processing step because the model may produce ill-formed outputs, for example, graphs with mismatched parentheses or disconnected components. To address this, a post-processing step is applied to correct formatting errors and reconstruct a well-formed AMR graph from its linearized representation. These steps are described in more detail in the following sections. 

\subsection{Experimental Setup}
To train a sequence-to-sequence AMR parser, we employ a multilingual language model mBart \citep{liu-etal-2020-multilingual-denoising}. To linearize AMR graph, we traverse the graph with depth first search (DFS) in line with \citet{Bevilacqua_Blloshmi_Navigli_2021}. As a pre-processing step, we rename variables in AMR graphs so that variable numbering follows an order (\textit{e.g.,} a, a2, a3$\cdots$) instead of random numbering (\textit{e.g.,} a3, a, a2$\cdots$). In addition, we added empty space between parentheses (see Figure \ref{fig:beforepp} and \ref{fig:afterpp} for differences between before and after pre-preprocessing). 

\begin{figure}[H]
\centering
\begin{minipage}{.45\textwidth}
  \centering
  \raggedright
  {\scriptsize\texttt{(m2 / multi-sentence \\
\qquad:snt1 (e / exact) \\
\qquad:snt2 (m / make-05 \\
\qquad\qquad:ARG2 (c1 / settlement \\
\qquad\qquad\qquad:ARG1-of (b / build-01 \\
\qquad\qquad\qquad\qquad:ARG0 (y / you))) \\
\qquad\qquad:ARG1 (p / point :quant 1)) \\
\qquad:snt3 (m1 / make-05 \\
\qquad\qquad:ARG2 (c2 / city) \\
\qquad\qquad:ARG1 (p1 / point :quant 2)))}}\vspace{7pt}
  \captionof{figure}{AMR graph before pre-processing.}
  \label{fig:beforepp}
\end{minipage}%
\hfill
\begin{minipage}{.45\textwidth}
  \centering
  \raggedright\vspace{18pt}
  {\scriptsize\texttt{( m / multi-sentence \\
\qquad:snt1 ( e / exact ) \\
\qquad:snt2 ( m2 / make-05 \\
\qquad\qquad:ARG2 ( s / settlement \\
\qquad\qquad\qquad:ARG1-of ( b / build-01 \\
\qquad\qquad\qquad\qquad:ARG0 ( y / you ) ) ) \\
\qquad\qquad\qquad:ARG1 ( p / point :quant 1 ) ) \\
\qquad:snt3 ( m3 / make-05 \\
\qquad\qquad:ARG2 ( c / city ) \\
\qquad\qquad:ARG1 ( p2 / point :quant 2 ) ) )}}
  \captionof{figure}{AMR graph after pre-processing.}
  \label{fig:afterpp}
\end{minipage}
\end{figure}

We train two distinct models: one trained solely on our data (hereafter referred to as Domain-specific), and another that is first trained on a larger AMR corpus \citep{knight-etal-2020-ldc2020} and then fine-tuned on our data (hereafter referred to as Pre-trained+Domain-specific). The aim of the second model is to explore whether leveraging large-scale AMR data can facilitate learning our data, which differs in several key aspects: data types (text vs.\ dialogue transcripts), domain (general vs.\ board game-related), and semantic roles (standard AMR vs.\ AMR adapted for French dialogue). Note that the current large-scale AMR data is only available in English and not in our target language, French. To obtain such data in French, we translated English AMR 3.0 into French using machine translation\footnote{\url{https://www.deepl.com/fr/translator}} following \citet{damonte-cohen-2018-cross}.  

We split our data set into train, dev and test sets to respectively train the model, to select the best checkpoint, and to evaluate the model's performance on unseen data. The training and dev set respectively consists of 1,375 and 146 examples.\footnote{We filtered out examples that include only non-annotable sound \textit{e.g.,} [\textit{rire}] and [\textit{toux}] - [laugh] and [cough] in English.} For testing, we used the subset of data that underwent a conflict resolution (see Section \ref{sec:corpus}), which consists of 146 examples after filtering out examples solely consisting of non-annotable words. 

The model was trained for 4,000 steps, with evaluations conducted every 50 steps on a dev set to select the best-performing checkpoint. Early stopping was applied, terminating training if the validation score did not improve over 750 consecutive steps. The learning rate was set to $3\mathrm{e}{-5}$. Pre-trained+Domain-specific was initially pre-trained on AMR 3.0 data for up to 40,000 steps, with early stopping triggered after 7,500 steps without improvement. Following pre-training, the model was fine-tuned on our data for 4,000 steps using the same settings described above for the Domain-specific training.

\subsection{Results}

Figure \ref{tab:results} shows the results of our experiments. The findings indicate that pre-training the model on large-scale data is beneficial to learn our corpus in several ways. First, it helps to learn the correct structure of AMR graphs. For example, while the Domain-specific model produced 3 ill-formed graphs out of 146 that could not be recovered during post-processing, the Pre-trained+Domain-specific model successfully avoided such errors.

Moreover, large-scale pre-training helps the model better identify the appropriate predicates for French text. The Domain-specific model occasionally produced predicates that closely resembled the surface form of the French verb, rather than the correct PropBank predicate. For instance, it generated \texttt{poser-01} instead of \texttt{put-01} for the phrase «tu peux poser...» (you can put...), and \texttt{peux-01} instead of \texttt{capable-01} for «tu peux » (you can).

\begin{table}[H]
\resizebox{\columnwidth}{!}{
\begin{tabular}{lr}
\toprule
                              & \textsc{smatch} \\
\midrule
Domain-specific               & 68.1            \\
Pre-trained+Domain-specific   & 73.5            \\ 
\bottomrule
\end{tabular}}
\caption{\textsc{smatch} scores of the two models.}
\label{tab:results}
\end{table}

Despite these improvements, both models exhibited certain weaknesses. Some sentences in the dataset included non-annotable elements such as coughing or laughter, marked with square brackets (\textit{e.g.,} [toux] for coughing, [rire] for laughing). These elements should not be represented in AMR graphs, but our model failed to capture the pattern and incorrectly annotated some of them (see Figures \ref{fig:gold} and \ref{fig:pred1} for an example). Additionally, although the Pre-trained+Domain-specific model generally performed better at predicting PropBank predicates for French verbs, both models struggled with rare verbs. In such cases, they generated incorrect predicates resembling the verb’s surface form—for example, \texttt{confine-01} instead of \texttt{entrust-01} for «on te confie...» (we entrust you with...).

\begin{figure}[H]
\centering
\begin{minipage}{.45\textwidth}
  \centering
  \raggedright
  {\footnotesize\texttt{(y / yes \\
\qquad:mod (a / ah))}}
  \captionof{figure}{Reference graph for « ah \textbf{[pron fin de mot fricative palatele sourde]}+ oui (0.5s) +[pron]» (ah [pronounce voiceless palatal fricative]+ yes (0.5s)).}
  \label{fig:gold}
\end{minipage}%
\hfill
\begin{minipage}{.45\textwidth}
  \centering
  \raggedright\vspace{11pt}
  {\footnotesize\texttt{
  (m / multi-sentence \\
\qquad:snt1 (a / ah) \\
\qquad:snt2 \textbf{(e / end-01} \\
\qquad\qquad:ARG1 (\textbf{w / word} \\
\qquad\qquad\qquad:mod \textbf{(f / fricative)}) \\
\qquad\qquad:ARG2 (y / yes)) \\
\qquad:snt3 (a2 / and \\
\qquad\qquad:op1 (y2 / yes)))}}
  \captionof{figure}{Pre-trained+Domain-specific's prediction for Figure \ref{fig:gold}.}
  \label{fig:pred1}
\end{minipage}
\end{figure}

Lastly, concerning new semantic roles added in our adaptation (\texttt{:discourse-marker} and \texttt{:back-channel}), both models showed good performance at capturing them. Among 43 discourse markers to predict, both models found around 30 discourse-markers (recall around~0.7). However, some of these discourse-markers were attached to wrong parent nodes. As for \texttt{:back-channel}, there was only one example in the test set and both models correctly predicted the \texttt{:back-channel}.

\section{Conclusion and Future Work}
We presented our ongoing work to annotate the DinG corpus in AMR to contribute to linguistic resources for French.
To better represent the dynamics of spontaneous speech in the DinG corpus, we adapted standard AMR by introducing new semantic roles. We provide an annotation guideline detailing these adaptations, as well as a data statement containing metadata of \corpus{}.\footnote{The annotation guideline, the data statement, and the corpus are available at \url{ https://doi.org/10.5281/zenodo.15537425}.} To demonstrate a practical application of the dataset, we trained and evaluated an AMR parser on our data. The resulting model can also serve as an annotation assistance tool, helping to accelerate the annotation process and scale up the semantic annotation process.
In our future work, we aim to expand the annotated dataset to approximately~3,000 utterances.

\paragraph{UMR}
Uniform Meaning Representation (UMR) have been introduced in \citet{van2021designing} as an extension of AMR to languages other than English, with the ambition of being used to ``annotate the semantic content of a text in any language''. UMR is developed as AMR with additional features, notably aspect, tense, modality, along with expanded ones, such as quantification \& scope, and discourse relations. 

While UMR appears as a very promising representation tool, we have not yet used it for our purposes. There is no French-UMR dataset available for now, which makes evaluation difficult, especially for corpora with complex language phenomena such as DinG. We plan to participate in the development of AMR to UMR translation tools, which should result in several silver French-UMR corpora, paving the way for further meaning representation work. The additions we made to AMR in order to annotate DinG are a lighter version of some of the additional annotations needed for UMR annotation; thus our annotation guidelines could also be of use for a middle step between AMR and UMR.

\section*{Acknowledgments}

We thank reviewers for their comments and suggestions.
We gratefully acknowledge the support of Institut Carnot Cognition (project ANAGRAM) and of the French National Research Agency (grant ANR-23-CE23-0017-01, project SynPaX).

\bibliographystyle{acl_natbib}
\bibliography{acl2021}

\end{document}